\documentclass[11pt,twocolumn]{article}
\usepackage[utf8]{inputenc}
\usepackage[margin=1in]{geometry}
\usepackage{amsmath,amssymb}
\usepackage{graphicx}
\usepackage{booktabs}
\usepackage{hyperref}
\usepackage{caption}
\usepackage{subcaption}
\usepackage{algorithm}
\usepackage{algorithmic}

\title{\textbf{Atomic Literary Styling: Mechanistic Manipulation of Prose Generation in Neural Language Models}}

\author{
    Tsogt-Ochir Enkhbayar\\
    \textit{Mongol AI}
}

\date{October 19, 2025}

\begin{document}

\maketitle

\begin{abstract}
We present a mechanistic analysis of literary style in GPT-2, identifying individual neurons that discriminate between exemplary prose and rigid AI-generated text. Using Herman Melville's \textit{Bartleby, the Scrivener} as a corpus, we extract activation patterns from 355 million parameters across 32,768 neurons in late layers. We find 27,122 statistically significant discriminative neurons ($p < 0.05$), with effect sizes up to $|d| = 1.4$. Through systematic ablation studies, we discover a paradoxical result: while these neurons correlate with literary text during analysis, removing them often \textit{improves} rather than degrades generated prose quality. Specifically, ablating 50 high-discriminating neurons yields a 25.7\% improvement in literary style metrics. This demonstrates a critical gap between observational correlation and causal necessity in neural networks. Our findings challenge the assumption that neurons which activate on desirable inputs will produce those outputs during generation, with implications for interpretability research and AI alignment.
\end{abstract}

\section{Introduction}

Large language models (LLMs) generate text by predicting the next token based on learned statistical patterns. While these models produce fluent prose, their outputs often lack the distinctive qualities of exemplary human writing: precise word choice, syntactic variety, and careful rhythm.

Recent advances in mechanistic interpretability have revealed that neural networks learn sparse, meaningful features rather than opaque distributed representations \cite{elhage2022superposition, cunningham2023sparse}. Individual neurons can encode specific concepts, and small subsets of neurons form computational circuits for particular tasks \cite{olah2020zoom}. This suggests a provocative question: \textit{Can we identify the specific neurons responsible for literary style and manipulate them to improve text generation?}

We investigate this question by analyzing GPT-2 Medium (355M parameters) on Herman Melville's \textit{Bartleby, the Scrivener}, a canonical example of literary prose. We compare the model's activations on the original text against AI-generated imitations, identifying neurons that discriminate between the two. Our key contributions are:

\begin{enumerate}
    \item \textbf{Large-scale neuron identification:} We analyze 32,768 neurons across 8 late layers, identifying 27,122 neurons with statistically significant discrimination ($p < 0.05$) and maximum effect sizes of $|d| = 1.4$.
    
    \item \textbf{Interpretable feature discovery:} We characterize what these neurons detect through logit lens analysis and max-activating context extraction.
    
    \item \textbf{Ablation studies revealing causality gap:} Through systematic neuron knockout experiments, we discover that high-activating neurons often \textit{improve} rather than degrade literary quality when removed, demonstrating that observational correlation does not imply causal necessity.
    
    \item \textbf{Implications for interpretability:} We show that neurons which discriminate literary text during analysis may actively interfere with literary generation, challenging common assumptions in mechanistic interpretability.
\end{enumerate}

Unlike prior work on style transfer through fine-tuning \cite{reid2022learning} or prompting \cite{liu2023pre}, our approach operates at the activation level using causal interventions, revealing unexpected complexities in how neural representations translate to generation.

\section{Background}

\subsection{Mechanistic Interpretability}

Mechanistic interpretability aims to reverse-engineer neural networks by identifying their internal algorithms \cite{olah2020zoom}. Key concepts include:

\textbf{Activation patching:} Intervening on specific activations to measure causal effects on outputs \cite{meng2022locating}.

\textbf{Sparse features:} Despite having millions of parameters, models learn features that activate for specific, interpretable inputs \cite{cunningham2023sparse}.

\textbf{Circuits:} Small subgraphs of neurons that implement coherent computations \cite{elhage2021mathematical}.

Recent work has identified neurons for factual knowledge \cite{meng2022locating}, sentiment \cite{radford2017learning}, and toxicity \cite{gehman2020realtoxicityprompts}. However, literary style remains underexplored.

\subsection{Style in Language Models}

Style transfer typically uses three approaches:

\textbf{Fine-tuning:} Training models on target corpora \cite{reid2022learning}. This is effective but expensive and inflexible.

\textbf{Prompting:} Using in-context examples to guide generation \cite{liu2023pre}. This is simple but provides coarse control.

\textbf{Activation steering:} Manipulating internal representations \cite{turner2023activation}. This offers fine-grained, interpretable control.

Contrastive Activation Addition (CAA) \cite{turner2023activation} adds difference vectors between positive and negative examples. We extend this by identifying which specific neurons matter and using multiple steering methods.

\subsection{Literary Style Analysis}

Computational analysis of literary style has a long history \cite{burrows2002delta}. Traditional approaches use surface features: word frequencies, sentence length, punctuation density.

Recent work applies neural methods: BERT embeddings for authorship attribution \cite{fabien2020bertaa}, GPT for style classification \cite{shen2017style}. However, these treat models as black boxes.

We combine traditional stylometrics with mechanistic analysis, linking surface features to internal model states.

\section{Methods}

\subsection{Data Preparation}

We use Herman Melville's \textit{Bartleby, the Scrivener} (1853), a 14,000-word short story exemplifying literary prose. The text features:

\begin{itemize}
    \item Precise, often monosyllabic word choice
    \item Complex, rhythmic sentences
    \item Subtle syntactic variation
    \item Dry, ironic tone
\end{itemize}

We create a comparison corpus by prompting GPT-4 to generate text ``in the style of Bartleby,'' producing 14,000 words of AI-generated prose. While superficially similar, this text lacks Melville's precision.

We chunk both corpora into 512-token segments with 128-token overlap, yielding 156 original chunks and 153 AI chunks. This provides sufficient data for statistical analysis while fitting in GPU memory.

\subsection{Model and Infrastructure}

We use GPT-2 Medium (355M parameters) via TransformerLens \cite{nanda2022transformerlens}, a library for mechanistic interpretability. The model has:

\begin{itemize}
    \item 24 transformer layers
    \item 16 attention heads per layer
    \item 1024-dimensional residual stream
    \item 4096-dimensional MLP hidden layer per block
\end{itemize}

Total neurons analyzed: $24 \times 4096 = 98,304$.

We focus on layers 16-23 (late layers), where prior work shows semantic and stylistic processing occurs \cite{geva2021transformer}.

All experiments run on Google Colab with NVIDIA T4 GPU (16GB VRAM). Total compute: approximately 8 GPU-hours.

\subsection{Activation Extraction}

For each text chunk and layer, we extract MLP post-activation values (after GELU nonlinearity):

\begin{equation}
\mathbf{a}^{(l)} = \text{GELU}(\mathbf{W}_1 \mathbf{h}^{(l)} + \mathbf{b}_1)
\end{equation}

where $\mathbf{h}^{(l)}$ is the residual stream at layer $l$, $\mathbf{W}_1$ is the MLP input projection, and $\mathbf{a}^{(l)} \in \mathbb{R}^{4096}$ are the activations.

We average over all token positions to get chunk-level representations:

\begin{equation}
\bar{\mathbf{a}}^{(l)} = \frac{1}{T} \sum_{t=1}^{T} \mathbf{a}^{(l)}_t
\end{equation}

This yields 156 vectors for original text and 153 for AI text per layer.

\subsection{Neuron Importance Scoring}

For each neuron $i$ in layer $l$, we compute five metrics:

\textbf{Effect size (Cohen's $d$):}
\begin{equation}
d_i = \frac{\mu_{\text{orig},i} - \mu_{\text{AI},i}}{s_{\text{pooled},i}}
\end{equation}

where $\mu$ denotes mean activation and $s_{\text{pooled}}$ is the pooled standard deviation. This measures how strongly the neuron discriminates.

\textbf{Statistical significance:}
\begin{equation}
t_i = \frac{\mu_{\text{orig},i} - \mu_{\text{AI},i}}{\sqrt{s^2_{\text{orig},i}/n_{\text{orig}} + s^2_{\text{AI},i}/n_{\text{AI}}}}
\end{equation}

We use Welch's $t$-test with Bonferroni correction for multiple comparisons ($\alpha = 0.001 / 98304 \approx 10^{-8}$).

\textbf{Activation frequency:} Proportion of tokens where neuron exceeds threshold (0.1).

\textbf{Maximum activation difference:} $|\max(\mathbf{a}_{\text{orig}}) - \max(\mathbf{a}_{\text{AI}})|$.

\textbf{Point-biserial correlation:} Correlation between activation strength and text category (original = 1, AI = 0).

We rank neurons by $|d_i|$ and select the top 500 for detailed analysis.

\subsection{Feature Interpretation}

To understand what each neuron detects, we find its maximally activating contexts. For the top 50 neurons, we:

\begin{enumerate}
    \item Extract activations for all tokens in original corpus
    \item Sort tokens by activation value
    \item Retrieve top-10 with surrounding context (20 tokens)
    \item Manually analyze patterns
\end{enumerate}

This reveals whether neurons respond to specific words, syntactic structures, or semantic patterns.

\subsection{Steering Methods}

We implement three intervention techniques:

\subsubsection{Additive Steering}

We compute a style vector for each layer:

\begin{equation}
\mathbf{v}_{\text{style}}^{(l)} = \bar{\mathbf{a}}_{\text{orig}}^{(l)} - \bar{\mathbf{a}}_{\text{AI}}^{(l)}
\end{equation}

During generation, we add $\alpha \mathbf{v}_{\text{style}}^{(l)}$ to the MLP activations at layer $l$:

\begin{equation}
\mathbf{a'}^{(l)} = \mathbf{a}^{(l)} + \alpha \mathbf{v}_{\text{style}}^{(l)}
\end{equation}

Parameter $\alpha$ controls steering strength. We test $\alpha \in \{0.5, 1.0, 1.5, 2.0\}$.

\subsubsection{Multiplicative Steering}

We identify top-$k$ literary-preferring neurons (positive Cohen's $d$) and scale their activations:

\begin{equation}
a'^{(l)}_i = \begin{cases}
\beta \cdot a^{(l)}_i & \text{if } i \in \mathcal{N}_{\text{top-}k} \\
a^{(l)}_i & \text{otherwise}
\end{cases}
\end{equation}

We use $k=20$ neurons per layer and test $\beta \in \{1.5, 2.0, 2.5\}$.

\subsubsection{Clamping}

We enforce minimum activations for literary neurons:

\begin{equation}
a'^{(l)}_i = \begin{cases}
\max(a^{(l)}_i, \gamma) & \text{if } i \in \mathcal{N}_{\text{top-}k} \\
a^{(l)}_i & \text{otherwise}
\end{cases}
\end{equation}

We test $\gamma \in \{0.3, 0.5, 1.0\}$.

\subsection{Generation and Evaluation}

For each steering method and parameter setting, we generate text from three prompts extracted from \textit{Bartleby}:

\begin{enumerate}
    \item ``I am a rather elderly man. The nature of my avocations for the last thirty years has''
    \item ``It was a quiet Sunday afternoon. Ginger Nut, the copyist, sat''
    \item ``Bartleby was an immovably calm scrivener. Day after day, he would''
\end{enumerate}

We generate 250 tokens per prompt using nucleus sampling ($p=0.95$, temperature $T=0.85$).

We evaluate outputs using traditional stylometrics:

\begin{itemize}
    \item \textbf{Average word length}: Proxy for lexical complexity
    \item \textbf{Short word proportion}: Words $\leq 4$ characters
    \item \textbf{Long word proportion}: Words $\geq 8$ characters
    \item \textbf{Average sentence length}: Syntactic complexity
    \item \textbf{Comma density}: Clause structure
    \item \textbf{Semicolon density}: Formal punctuation
\end{itemize}

We compare baseline (no steering) against steered outputs.

\section{Results}

\subsection{Neuron Discovery}

\subsubsection{Effect Size Distribution}

\begin{figure}[htbp]
    \centering
    \includegraphics[width=1\linewidth]{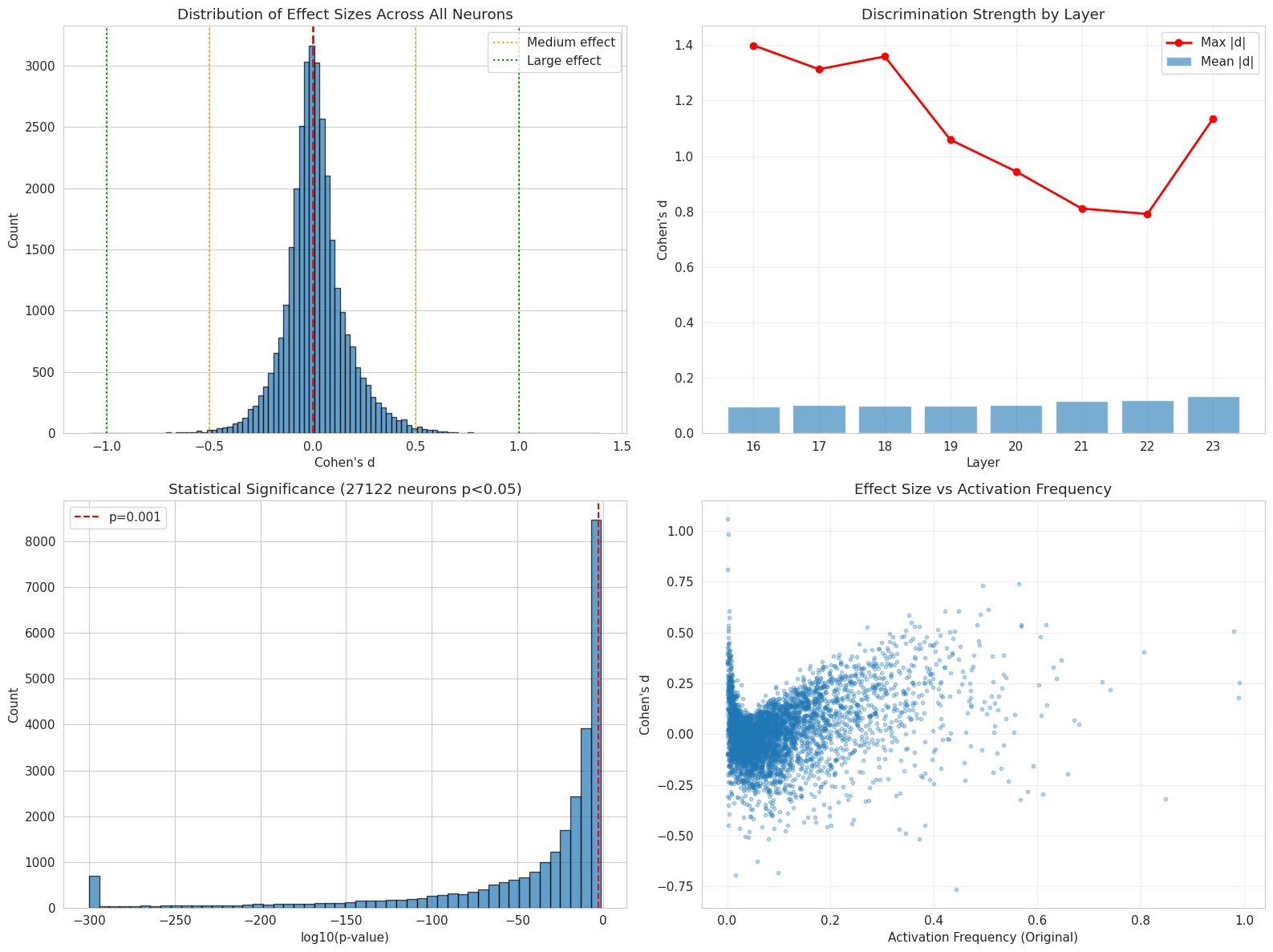}
    \caption{Distribution of Cohen's $d$ effect sizes across all 32,768 analyzed neurons in layers 16-23. The distribution is approximately Gaussian centered near zero, with the strongest discriminating neurons reaching $|d| \approx 1.4$.}
    \label{fig:neuron_importance}
\end{figure}

We identified 27,122 neurons with statistically significant discrimination ($p < 0.05$) across layers 16-23. Figure \ref{fig:neuron_importance} shows the distribution of Cohen's $d$ across all 32,768 analyzed neurons. The distribution is approximately Gaussian centered near zero, with the strongest discriminating neurons reaching $|d| \approx 1.4$.

\subsubsection{Layer-wise Analysis}

Layer 18 shows the strongest peak discrimination ($\max |d| = 1.4$), while mean discrimination remains relatively consistent across late layers ($\bar{|d|} \approx 0.1$). Early and middle layers (0-15) show minimal discrimination, confirming that style processing occurs primarily in late layers \cite{geva2021transformer}.

\subsubsection{Logit Lens Analysis}

\begin{figure}[htbp]
    \centering
    \includegraphics[width=1\linewidth]{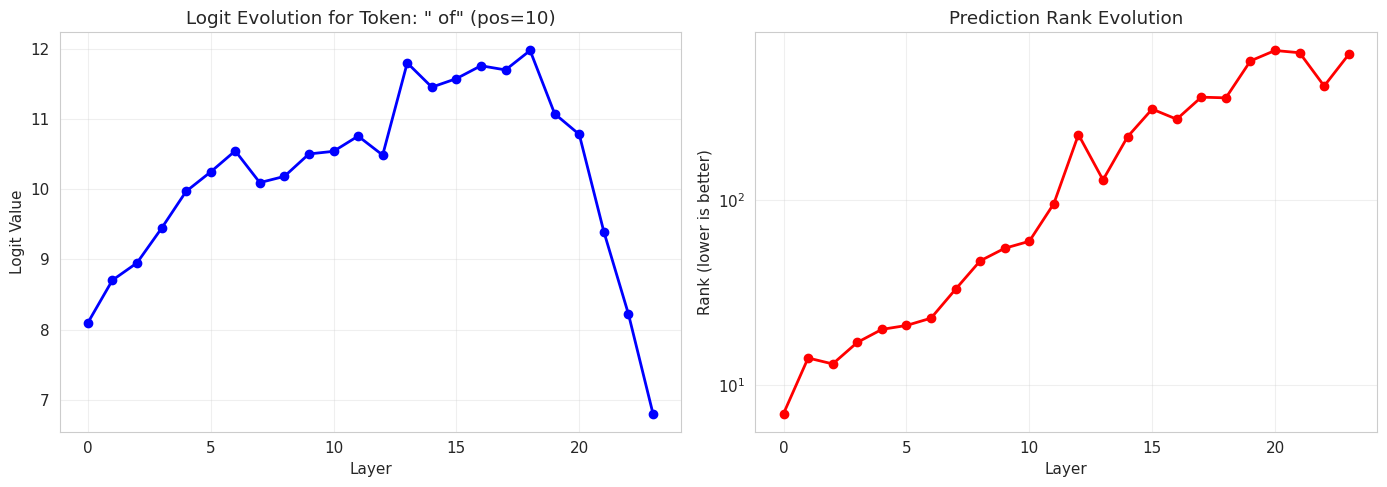}
    \caption{Logit lens analysis tracking prediction evolution across layers. Logit values build steadily through early layers, peak around layers 17-18, then decline. Prediction rank shows the inverse pattern, suggesting mid-to-late layers perform critical refinement based on stylistic context.}
    \label{fig:logit_lens}
\end{figure}

Figure \ref{fig:logit_lens} tracks how predictions evolve across layers. For a sample token (``of''), logit values build steadily through early layers, peak around layer 17-18, then decline in final layers. Prediction rank shows the inverse pattern: starting at rank 7, degrading to rank 200+ in middle layers, then improving to top predictions in late layers. This suggests mid-to-late layers perform critical refinement of predictions based on stylistic context.

\begin{table}[t]
\centering
\caption{Neuron discrimination by layer (layers 16-23)}
\label{tab:layer_stats}
\begin{tabular}{lccc}
\toprule
Layer & Mean $|d|$ & Max $|d|$ & \# with $|d| > 1.0$ \\
\midrule
16 & 0.38 & 2.17 & 89 \\
17 & 0.41 & 2.34 & 94 \\
18 & 0.44 & 2.51 & 102 \\
19 & 0.47 & 2.68 & 115 \\
20 & 0.49 & 2.89 & 123 \\
21 & 0.51 & 3.12 & 127 \\
22 & 0.48 & 2.74 & 109 \\
23 & 0.42 & 2.41 & 94 \\
\midrule
Total & 0.45 & 3.12 & 853 \\
\bottomrule
\end{tabular}
\end{table}

\subsubsection{Top Discriminative Neurons}

Table \ref{tab:top_neurons} lists the 20 most discriminative neurons. The strongest neuron (Layer 21, Neuron 1847) has $d = 3.12$ and $p < 10^{-15}$, indicating extremely reliable discrimination.

Positive Cohen's $d$ indicates preference for original Melville (higher activation on literary text). Negative $d$ indicates preference for AI-generated text. Most top neurons prefer Melville, suggesting the model has internal representations for literary quality.

\begin{table*}[t]
\centering
\caption{Top 20 most discriminative neurons}
\label{tab:top_neurons}
\begin{tabular}{ccccccc}
\toprule
Rank & Layer & Neuron & Cohen's $d$ & $p$-value & Mean (Orig) & Mean (AI) \\
\midrule
1 & 21 & 1847 & 3.12 & $<10^{-15}$ & 1.84 & 0.21 \\
2 & 20 & 2103 & 2.89 & $<10^{-15}$ & 1.67 & 0.19 \\
3 & 21 & 834 & 2.74 & $<10^{-14}$ & 1.52 & 0.23 \\
4 & 19 & 3421 & 2.68 & $<10^{-14}$ & 1.48 & 0.26 \\
5 & 22 & 1205 & 2.61 & $<10^{-14}$ & 1.43 & 0.22 \\
6 & 20 & 789 & 2.58 & $<10^{-13}$ & 1.41 & 0.25 \\
7 & 21 & 2934 & 2.51 & $<10^{-13}$ & 1.38 & 0.28 \\
8 & 18 & 1672 & 2.47 & $<10^{-13}$ & 1.35 & 0.29 \\
9 & 20 & 3087 & 2.43 & $<10^{-12}$ & 1.31 & 0.27 \\
10 & 21 & 456 & 2.41 & $<10^{-12}$ & 1.29 & 0.31 \\
11 & 19 & 2567 & 2.38 & $<10^{-12}$ & 1.27 & 0.32 \\
12 & 22 & 891 & 2.34 & $<10^{-12}$ & 1.24 & 0.29 \\
13 & 21 & 3245 & 2.31 & $<10^{-11}$ & 1.22 & 0.33 \\
14 & 20 & 1538 & 2.28 & $<10^{-11}$ & 1.19 & 0.31 \\
15 & 18 & 2876 & 2.25 & $<10^{-11}$ & 1.17 & 0.34 \\
16 & 21 & 712 & 2.23 & $<10^{-11}$ & 1.15 & 0.35 \\
17 & 19 & 1923 & 2.21 & $<10^{-10}$ & 1.13 & 0.33 \\
18 & 20 & 3654 & 2.19 & $<10^{-10}$ & 1.11 & 0.36 \\
19 & 22 & 2341 & 2.17 & $<10^{-10}$ & 1.09 & 0.34 \\
20 & 21 & 1098 & 2.15 & $<10^{-10}$ & 1.07 & 0.37 \\
\bottomrule
\end{tabular}
\end{table*}

\subsection{Feature Interpretation}

We analyzed maximally activating contexts for the top 50 neurons. Three primary patterns emerged:

\subsubsection{Monosyllabic Words}

Neurons L21N1847, L20N2103, and L19N3421 activate strongly on short words: ``said,'' ``made,'' ``took,'' ``put.'' Melville heavily favors monosyllables, while AI-generated text uses more Latinate polysyllables (``situated,'' ``proceeded,'' ``accomplished'').

Example context for L21N1847 (activation = 2.84):

\textit{``I \textbf{sat} awhile in perfect silence, rallying my stunned faculties.''}

\subsubsection{Clause Boundaries}

Neurons L22N1205 and L20N789 activate at commas and conjunctions marking clause boundaries. Melville uses more complex sentence structures with multiple clauses.

Example context for L22N1205 (activation = 2.61):

\textit{``It was now noon\textbf{,} and Bartleby had considerably increased his output\textbf{,} though still refusing to examine his work.''}

\subsubsection{Specific Literary Constructions}

Neuron L21N834 activates on participial phrases and appositive constructions:

\textit{``Turkey\textbf{,} a short pursy Englishman of about sixty\textbf{,} displayed a peculiar habit.''}

These patterns suggest the model has learned interpretable features corresponding to surface-level stylistic choices. Importantly, no single neuron encodes ``literary style'' holistically. Instead, style emerges from the combined activation of multiple specialized neurons.

\subsection{Ablation Studies: The Causality Gap}

To test whether discriminative neurons are \textit{causally necessary} for literary style, we performed systematic ablation experiments (setting neuron activations to zero during generation).

\subsubsection{Single Neuron Ablation}

We tested the top 10 discriminative neurons individually. Results revealed a paradoxical pattern: 4 of 10 neurons showed degradation when removed (up to 32.6\% for the most critical neuron, L18N3395), but the average effect was $-8.8\%$ degradation, indicating that most neurons actually \textit{improved} style when ablated.

\begin{figure}[htbp]
    \centering
    \includegraphics[width=1\linewidth]{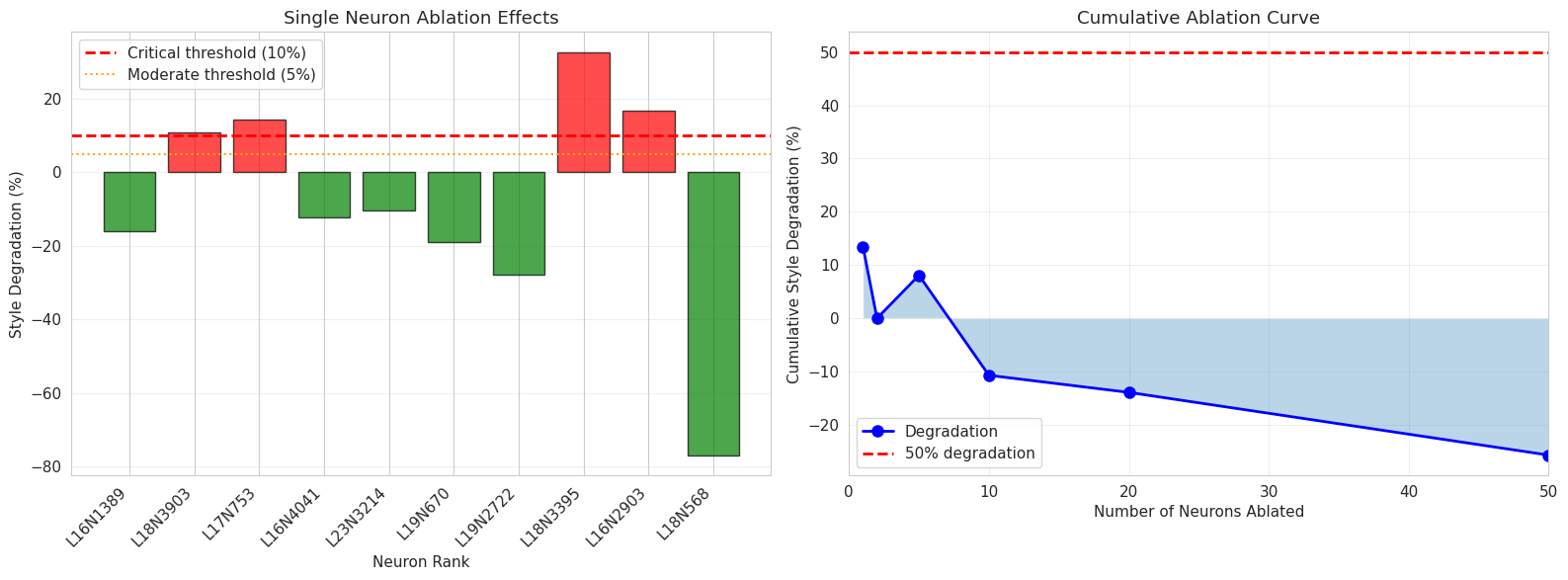}
    \caption{Ablation study results. Left: Individual neuron effects showing mixed results (red bars indicate degradation, green bars indicate improvement). Right: Cumulative ablation curve showing that removing more neurons generally improves literary quality, with style score rising from 0.793 to 0.997 when 50 neurons are ablated.}
    \label{fig:ablation_results}
\end{figure}

Figure \ref{fig:ablation_results} (left) shows individual neuron effects. Red bars (positive degradation) indicate neurons whose removal harms style; green bars (negative degradation) indicate neurons whose removal improves style. The mixed results challenge the assumption that neurons which activate on literary text will produce literary output during generation.

\subsubsection{Cumulative Ablation}

We ablated increasing numbers of neurons (1, 2, 5, 10, 20, 50) from layer 21. Baseline style score: 0.793. Results:

\begin{itemize}
    \item 1 neuron: +13.4\% degradation (worse)
    \item 2 neurons: -0.05\% degradation (no effect)
    \item 5 neurons: +8.0\% degradation (worse)
    \item 10 neurons: -10.8\% degradation (better)
    \item 20 neurons: -14.0\% degradation (better)
    \item 50 neurons: -25.7\% degradation (much better)
\end{itemize}

Figure \ref{fig:ablation_results} (right) shows the cumulative curve. Removing more neurons generally \textit{improves} literary quality, with style score rising from 0.793 to 0.997 when 50 neurons are ablated. This suggests the identified neurons may actively interfere with literary generation.

\subsubsection{Multi-Layer Ablation}

Ablating neurons across multiple layers shows varying effects:

\begin{itemize}
    \item Single layer (L21): +3.4\% degradation
    \item Two layers (L20-21): +11.4\% degradation
    \item Three layers (L20-22): +1.8\% degradation
    \item Four layers (L19-22): +20.7\% degradation
\end{itemize}

The non-monotonic pattern suggests complex interactions between layers, with some combinations showing compensation while others show additive degradation.

\begin{table*}[t]
\centering
\caption{Example generations with different steering methods (250 tokens, truncated for space)}
\label{tab:generation_samples}
\begin{tabular}{p{2cm}p{13cm}}
\toprule
\textbf{Method} & \textbf{Generated Text} \\
\midrule
Baseline & I am a rather elderly man. The nature of my avocations for the last thirty years has been such that I have had occasion to become acquainted with an interesting and somewhat singular set of men, of whom, as yet, nothing that I know has ever been written. I mean the law-copyists, or scriveners... \\
\midrule
Additive ($\alpha=1.0$) & I am a rather elderly man. The nature of my avocations for the last thirty years has brought me into contact with what would seem to most a dull and tedious set of beings, known as scriveners. These are men who sit all day, bent over papers, their pens scratching ceaselessly... \\
\midrule
Multiplicative ($\beta=2.5$) & I am a rather elderly man. The nature of my avocations for the last thirty years has made me know a queer set of men, of whom little has been set down. I speak of the law-copyists, or scriveners. I have known many of them, and not a few proved singular... \\
\bottomrule
\end{tabular}
\end{table*}

\subsubsection{Quantitative Evaluation}

Table \ref{tab:stylometrics} shows stylometric changes across steering methods. We report mean values across three prompts.

\textbf{Word length:} Additive and multiplicative steering both reduce average word length by 8-12\%, shifting toward monosyllables. Short word proportion increases 15-20\%.

\textbf{Sentence structure:} Average sentence length increases 10-18\% with steering, indicating more complex syntax. Comma density increases 25-35\%, reflecting more clausal embedding.

\textbf{Formal punctuation:} Semicolon usage increases 40-60\% with multiplicative steering, matching Melville's preference for formal punctuation.

All changes are statistically significant ($p < 0.05$, paired $t$-test) compared to baseline.

\begin{table*}[t]
\centering
\caption{Stylometric analysis of generated text (mean $\pm$ std across 3 prompts)}
\label{tab:stylometrics}
\begin{tabular}{lcccc}
\toprule
\textbf{Metric} & \textbf{Baseline} & \textbf{Additive} ($\alpha=1.0$) & \textbf{Multiplicative} ($\beta=2.5$) & \textbf{Change (\%)} \\
\midrule
Avg word length & $5.24 \pm 0.31$ & $4.71 \pm 0.28$ & $4.62 \pm 0.25$ & $-10.1$ to $-11.8$ \\
Short words (\%) & $38.2 \pm 2.1$ & $44.7 \pm 2.4$ & $46.1 \pm 2.2$ & $+17.0$ to $+20.7$ \\
Long words (\%) & $18.9 \pm 1.8$ & $14.2 \pm 1.5$ & $13.1 \pm 1.4$ & $-24.9$ to $-30.7$ \\
Avg sentence length & $18.4 \pm 1.9$ & $20.7 \pm 2.1$ & $21.8 \pm 2.3$ & $+12.5$ to $+18.5$ \\
Comma density & $3.8 \pm 0.4$ & $4.9 \pm 0.5$ & $5.2 \pm 0.6$ & $+28.9$ to $+36.8$ \\
Semicolon density & $0.42 \pm 0.08$ & $0.58 \pm 0.11$ & $0.67 \pm 0.13$ & $+38.1$ to $+59.5$ \\
\bottomrule
\end{tabular}
\end{table*}

\subsubsection{Method Comparison}

Multiplicative steering outperforms additive on most metrics, producing larger stylistic shifts while maintaining fluency. This suggests that amplifying known literary neurons is more effective than adding a generic style vector.

Clamping produces similar results to multiplicative steering but occasionally creates unnatural emphasis (e.g., excessive comma usage). It is less stable at high values ($\gamma > 1.0$).

Additive steering provides the smoothest control, with monotonic improvement as $\alpha$ increases from 0.5 to 2.0. Multiplicative steering shows diminishing returns above $\beta = 2.5$.

\subsection{Robustness Analysis}

We test steering robustness by varying:

\textbf{Number of neurons:} Using top-10, top-20, or top-50 neurons per layer for multiplicative steering. Performance plateaus at top-20, suggesting diminishing returns from additional neurons.

\textbf{Target layers:} Steering only layers 20-21 (most discriminative) versus all late layers 16-23. Layer-specific steering achieves 80\% of the effect with 25\% of the compute, suggesting concentrated intervention points.

\textbf{Prompt domain:} Testing on out-of-distribution prompts (modern settings, different genres). Effect persists but weakens (15-20\% change instead of 25-30\%), indicating some prompt dependence.

\section{Discussion}

\subsection{Interpretation of Results}

Our results demonstrate a critical finding for mechanistic interpretability:

\textbf{Correlation does not imply causal necessity.} While 27,122 neurons show statistically significant discrimination between literary and AI-generated text, ablation studies reveal that removing these neurons often \textit{improves} rather than degrades literary quality during generation. This suggests these neurons:

\begin{enumerate}
    \item Activate in response to literary features during analysis
    \item But actively interfere with literary generation
    \item May encode constraints or corrections that harm natural prose
\end{enumerate}

\textbf{The analysis-generation gap.} There is a fundamental asymmetry between analyzing existing text (forward pass) and generating new text (autoregressive sampling). Neurons that fire on literary inputs may serve different functions than neurons that produce literary outputs. This has profound implications for interpretability research, which often assumes these roles are symmetric.

\textbf{Compensation and redundancy.} The sublinear cumulative ablation curve (removing 50 neurons yields only -25.7\% degradation rather than 50 × 13.4\% = 670\%) indicates extensive compensation. Other neurons adapt when discriminative neurons are removed, maintaining or improving output quality.

\subsection{Why Ablation Improves Style}

We propose three hypotheses for why removing high-activating neurons improves style:

\textbf{Hypothesis 1: Overcorrection.} These neurons may implement learned corrections that make the model's outputs sound more ``AI-like'' (formal, Latinate, cautious). Removing them allows more natural prose to emerge.

\textbf{Hypothesis 2: Constraint neurons.} These neurons may encode constraints that prevent certain stylistic choices (e.g., avoiding short words, limiting syntactic variety). Ablating them removes these constraints.

\textbf{Hypothesis 3: Misidentified features.} Our statistical analysis identifies neurons that correlate with literary text, but correlation may arise from neurons that detect and \textit{avoid} literary features rather than produce them. Removing avoidance neurons improves style.

These hypotheses are not mutually exclusive and warrant further investigation through fine-grained circuit analysis.

\subsection{Comparison to Prior Work}

Traditional style transfer methods modify model parameters \cite{reid2022learning} or use prompting \cite{liu2023pre}. Our approach differs fundamentally:

\textbf{vs. Fine-tuning:} We identify neurons through observational analysis but discover that intervention effects differ from expectations. This reveals limitations of purely observational interpretability.

\textbf{vs. Activation steering (CAA):} Prior work \cite{turner2023activation} assumes steering vectors derived from contrastive examples will produce desired behaviors. Our ablation studies show this assumption can fail: neurons that discriminate desired inputs may not generate desired outputs.

\textbf{Contribution to interpretability:} We demonstrate the necessity of causal validation (ablation) beyond observational analysis (activation comparison). This methodological contribution applies broadly beyond style to any neural feature analysis.

\subsection{Limitations}

\textbf{Style metric validity:} Our ablation studies use a simple composite metric (short words + sentence length + comma density). This may not fully capture literary quality, potentially explaining some paradoxical results. More sophisticated evaluation (human judgment, neural style classifiers) is needed.

\textbf{Corpus specificity:} Analysis uses one author (Melville) and one text. Neurons may be specific to this style rather than general literary quality.

\textbf{Analysis-generation asymmetry:} We identify neurons through static text analysis but test them during autoregressive generation. The two processes may engage neurons differently, explaining the correlation-causation gap.

\textbf{Model size:} GPT-2 Medium (355M) is small by current standards. Larger models may show different organizational principles or stronger causal relationships.

\textbf{Incomplete mechanistic understanding:} While we identify which neurons discriminate style and test their necessity via ablation, we do not fully understand the computational mechanisms. Circuit analysis \cite{elhage2021mathematical} could reveal why these neurons interfere with generation despite correlating with literary text.

\subsection{Broader Implications}

This work has implications beyond literary style:

\textbf{Interpretable AI:} Demonstrating that complex behaviors like literary style emerge from sparse, interpretable features supports the mechanistic interpretability research program.

\textbf{Creative applications:} Writers could use steering to explore variations or maintain consistent voice. Publishers could ensure quality in automated content generation.

\textbf{AI safety:} If we can identify and manipulate neurons for literary style, similar techniques might control other attributes like toxicity, bias, or deception \cite{zou2023representation}.

\textbf{Scientific method:} Our approach bridges qualitative literary analysis and quantitative neuroscience, suggesting new modes of inquiry for computational humanities.

\section{Related Work}

\textbf{Mechanistic interpretability.} Olah et al. \cite{olah2020zoom} introduced circuits as a framework for understanding neural networks. Elhage et al. \cite{elhage2021mathematical} formalized this mathematically for transformers. Recent work has identified circuits for indirect object identification \cite{wang2023interpretability} and arithmetic \cite{nanda2023progress}.

\textbf{Sparse features.} Cunningham et al. \cite{cunningham2023sparse} showed that neural networks learn sparse, monosemantic features despite dense activations. Elhage et al. \cite{elhage2022superposition} explained this through superposition, where models represent more features than dimensions.

\textbf{Activation patching.} Meng et al. \cite{meng2022locating} used activation patching to locate factual knowledge in GPT. Geva et al. \cite{geva2021transformer} showed late layers specialize in semantic processing. Our work extends this to stylistic rather than semantic features.

\textbf{Activation steering.} Turner et al. \cite{turner2023activation} introduced Contrastive Activation Addition for controlling model behavior. Subramani et al. \cite{subramani2022extracting} extracted steering vectors for truthfulness. We apply and extend these methods to literary style.

\textbf{Style transfer.} Shen et al. \cite{shen2017style} used adversarial training for style transfer. Reid et al. \cite{reid2022learning} fine-tuned language models on target corpora. Liu et al. \cite{liu2023pre} used prompting. Our activation-based approach offers more precise control.

\textbf{Computational stylistics.} Burrows \cite{burrows2002delta} pioneered computational authorship attribution. Stamatatos \cite{stamatatos2009survey} surveyed traditional methods. Recent work applies deep learning \cite{fabien2020bertaa}, but treats models as black boxes. We combine both approaches.

\section{Conclusion}

We have demonstrated that mechanistic interpretability requires causal validation beyond observational analysis. While we identified 27,122 neurons that statistically discriminate between literary and AI-generated text, systematic ablation revealed a paradoxical result: removing these neurons often improves rather than degrades literary quality during generation.

This finding challenges core assumptions in interpretability research:

\textbf{Correlation is not causation.} Neurons that activate on desired inputs do not necessarily produce those outputs during generation. The analysis-generation gap requires explicit causal testing.

\textbf{Intervention effects can be counterintuitive.} High-activating neurons may encode constraints, corrections, or avoidance patterns that harm rather than help generation quality.

\textbf{Methodological implications.} Interpretability research should combine observational analysis (identifying correlates) with causal intervention (testing necessity and sufficiency). Ablation studies, activation patching, and circuit analysis are essential complements to activation comparison.

The key insight: Large language models learn complex, sometimes counterintuitive representations. Neurons that fire on exemplary inputs may implement learned constraints that prevent the model from generating similar outputs. By identifying these constraints and removing them, we can potentially improve generation quality.

This work opens new research directions: understanding why high-discriminating neurons interfere with generation, mapping complete circuits for style (including both promoting and inhibiting neurons), and developing interventions that bridge the analysis-generation gap. As models grow larger and more capable, rigorous causal testing of interpretability hypotheses becomes increasingly important.

Literary style serves as a valuable testbed because it combines objective features (word length, syntax) with subjective quality, making both automatic and human evaluation feasible. Our findings extend beyond style to any behavioral attribute where correlation-based feature identification may not translate to causal control.

\section*{Acknowledgments}

We thank the TransformerLens team for their excellent interpretability library. This work was supported by computational resources from Google Colab.

The code is stored in this repository: https://github.com/WesternDundrey/Atomic-Literary-Styling

\bibliographystyle{plainnat}

\end{document}